# Landmark Weighting for 3DMM Shape Fitting

Yu Yang[a], Xiao-Jun Wu[a, *] and Josef Kittler[b]

[a]School of Internet of Things Engineering, Jiangnan University, Wuxi 214122, China
[b]CVSSP, University of Surrey, Guildford, GU2 7XH, UK

## ABSTRACT

*Abstract: Human face is a 3D object with shape and surface texture. 3D Morphable Model (3DMM) is a powerful tool for reconstructing the 3D face from a single 2D face image. In the shape fitting process, 3DMM estimates the correspondence between 2D and 3D landmarks. Most traditional 3DMM fitting methods fail to reconstruct an accurate model because face shape fitting is a difficult non-linear optimization problem. In this paper we show that landmark weighting is instrumental to improve the accuracy of shape reconstruction and propose a novel 3D Morphable Model Fitting method. Different from previous works that treat all landmarks equally, we take into consideration the estimated errors for each pair of 2D and 3D corresponding landmarks. The landmark points are weighted in the optimization cost function based on these errors. Obviously, these landmarks have different semantics because they locate on different facial components. In the context of the solution of fitting is approximated, there are deviations in landmarks matching. However, these landmarks with different semantics have different effects on reconstructing 3D faces. Thus, it is necessary to consider each landmark individually. To our knowledge, we are the first to analyze each feature point for 3D face reconstruction by 3DMM. The weight is adaptive with the estimation residuals of landmarks. Experimental results show that the proposed method significantly reduces the reconstruction error and improves the authenticity of the 3D model expression.*

## 1. Introduction

3D face surface reconstruction is an important computer vision task. It arises in many applications, such as 3D-assisted face recognition [1], facial expression analysis [2,3], and 3D animation [4,5]. The 3D morphable face model(3DMM) proposed by Blanz and Vetter in 1999 [6], is a source of prior knowledge which can be used to reconstruct a specific 3D face including its pose and illumination from a given single 2D image. The original 3DMM fitting method reconstructs the face by means of analysis-synthesis framework whereby Gauss-Newton optimization is applied to minimize the difference between the input image and its synthesized.

Traditional fitting algorithms estimate the parameters of the morphable model to recover the pose and illumination condition for the 3D face from a given image. As the 3D Morphable Model contains many free parameters; namely facial shape and texture parameters, camera and lighting parameters, the fitting process is very time-consuming and suffers from the local minimum problem, just as other Gauss-Newton based methods. In order to improve the 3DMM fitting efficiency and accuracy, many innovative algorithms have been proposed in recent years. On the one hand, the new directions include adopting CNN [7] or regression operators [8,9,10] learnt from training samples to estimate the 3DMM parameters. On the other hand, sparse feature points or feature operators are used instead of using all the points on the 3D shape to recover 3DMM parameters [11]. In addition, Hu proposed U-3DMM [12] to accelerate 3DMM fitting in the presence of occlusion and noise.

None of the existing fitting methods take into account the landmarking error, Moreover, different feature points have different semantics, and this effect on the reconstruction errors is various. This is particularly important for landmark-based algorithms, the fitting process is optimized only at landmarks. Unfortunately, in many cases, the positions of these available landmark points are not accurate enough for 3D shape fitting. Accuracy is particularly important in expression defining areas (such as eyes, nose, mouth, eye brows). The traditional fitting methods may produce large fitting errors in these important feature regions and result in a bad reconstruction. Therefore, we consider to weight each landmark in the fitting optimization process, to prevent their undue influence. The residual weights and 3DMM parameters are alternatively optimized by an iterative process. The larger the error, the greater the weight. Our experimental results show that by virtue of weighting the global reconstruction error is reduced, the match of the landmarks points become more accurate, and the reconstructed face appears significantly better.

This paper is organized as follows: Section 2 presents the related work. In Section 3, we give a brief introduction to the 3DMM, Section 4 describes the details of our proposed method. Section 5 presents the results of an experimental validation of the proposed method. Section 6 concludes the paper.

* Corresponding author. Tel.: +86-13601487486; e-mail: wu_xiaojun@jiangnan.edu.cn

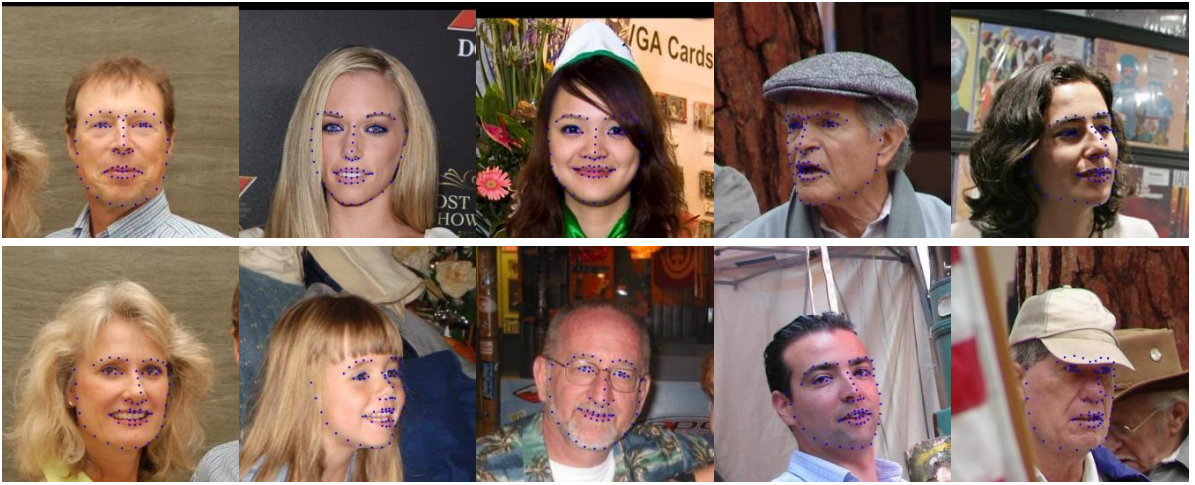

Figure 1. Faces with landmarks

## 2. Related Work

### 2.1. 3D Face Models

Reconstructing the 3D face surface from a set of input such as image(s), video and depth data is a longstanding problem in computer vision. In the biometrics field, the pose, illumination and expression are invariant for the 3D face model, and an accurate person-specific 3D face model may improve face recognition compared to a 2D face model. In graphics and animation community, the high-fidelity 3D face is desired. Blanz and Vetter proposed a 3D morphable model (3DMM) for modeling 3D human face from a single facial image.3DMM is established by projecting texture and shape from a training set of 3D face scans into the PCA space. A 3D face is represented as a linear combination of basis shape and texture vectors. To construct a 3D face from a 2D facial image, 3DMM-based methods estimate the texture and shape coefficients by minimizing the error between feature points which are on the input 2D facial image and corresponding projected points which are from the reconstructed 3D face. The traditional 3DMM is effective on estimating face pose and illumination variations, and has been applied to many subsequent works, such as 3D-assisted face recognition [13,14,15], face frontalization [16,17] and face alignment [10,18,19], etc.

With the higher performance of the scanning device and improved registration algorithms, the 3DMM was further refined based on large high-resolution scan dataset with a wide variety of age, gender, and ethnicity. The Baesl Face Model [20] and LSFM-3DMM [21] are two widely used models in recent years. In order to express the variation of face expression, the extended 3DMM [18,22,23,24] (E-3DMM) was suggested. Large number of 3D scans with diverse expressions have been used to train a 3D face model [25]. Recently, face alignment and 3D-assisted face recognition based on E-3DMM has been greatly improved in accuracy.

### 2.2. Fitting

Recovering the 3D model from a specific 2D facial image involves estimating the pose, shape, facial texture and illumination via a fitting process. To minimize the differences between 2D input image and its 3D reconstruction, the original method performs the fitting on all facial pixels based on the analysis-by-synthesis framework, which suffers from a high computational load and sensitive to initialization. Due to the computational complexity and challenges of optimization, it is difficult to achieve efficient and accurate results. Recently, many algorithms have been proposed to improve accuracy of the 3DMM fitting process.

Stochastic Newton Optimization (SNO) [26] randomly chooses a small subset from the entire face model represented by a mesh of vertices and optimizes the fitting cost function only on the subset. Although SNO reduces the computational complexity, the random selection of subset leads to an unstable fitting effect.

The Inverse Compositional Image Alignment (ICIA) [27,28] was extended to 3DMM fitting. It facilitates pre-computing the derivatives of the cost function to improve the efficiency of fitting. 2D landmarks are used directly for shape fitting in [29] as the accuracy of face landmarking methods increases. Multi-Feature Fitting (MFF) [30] extracts multiple features to construct a smooth objective function. Local features and cascaded regression were applied to shape fitting [31]. A recent work [12] modified the cost function to a unified linear framework to enhance the robustness of fitting to occlusion and achieved promising results.

## 3. Traditional 3D Morphable Model

The traditional 3D Morphable Model represents a 3D face with a shape-vector $S' = (X_1, Y_1, Z_1, X_2, Y_2, Z_2, \ldots, X_n, Y_n, Z_n)^T \in \Re^{3n}$ which contains n vertices in Cartesian coordinates and a texture-vector $T' = (R_1, G_1, B_1, R_2, G_2, B_2, \ldots, R_n, G_n, B_n)^T \in \Re^{3n}$ which contains the R, G, B color values of the n corresponding vertices. PCA is then applied to a set of exemplars faces. separately, to create the shape $S_{model}$ and texture $T_{model}$ model that can be described as:

$$S_{model} = \overline{S} + \sum_{i=1}^{m} \alpha_i s_i \quad T_{model} = \overline{T} + \sum_{i=1}^{m} \beta_i t_i \quad (1)$$

In (1) $\overline{S}$ and $\overline{T}$ are the mean shape and texture respectively, m is the number of eigenvectors of the shape and texture covariance matrices, $s_i$ and $t_i$ are the $i$th eigenvectors of shape and texture covariance matrices respectively, $\boldsymbol{\alpha} = (\alpha_1, \alpha_2, \ldots, \alpha_m)$ and $\boldsymbol{\beta} = (\beta_1, \beta_2, \ldots, \beta_m)$ are shape and texture parameters of $S_{model}$ and $T_{model}$, respectively.

In addition, expression variation can be added to 3DMM. Face expression can be seen as a linear combination of expression deviation from the neutral face:

$$S_{model} = \overline{S} + \sum_{i=1}^{m} \alpha_{i,id} s_{i,id} + \sum_{j=1}^{k} \alpha_{j,exp} s_{j,exp} \quad (2)$$

where $s_{i,id}$ is the $i$th eigenvector of neutral expression face, $\boldsymbol{\alpha_{id}} = (\alpha_{1,id}, \alpha_{2,id}, \ldots, \alpha_{m,id})^T$ is shape parameter, $s_{j,exp}$ is the $i$th eigenvector of the offset between expression scans and neutral scans, and $\boldsymbol{\alpha_{exp}} = (\alpha_{1,exp}, \alpha_{2,exp}, \ldots, \alpha_{k,exp})^T$ is the expression parameter vector.

In the fitting process, to project the 3DMM to 2D face space, we assume weak perspective camera projection:

$$S_{2d} = f\begin{bmatrix}1 & 0 & 0\\ 0 & 1 & 0\end{bmatrix}\mathcal{R}(S_{model} + t_{3d}) \quad (3)$$

where $S_{2d}$ is the vector of 2D coordinates of the registered 3D points of the 3D face model, $f$ is the scale, $\mathcal{R} \in \Re^{3n}$ is a rotation matrix which defines: pitch, yaw and roll directions of the 3D face and $t_{3d}$ is the 3D translation. The goal of 3DMM fitting is to minimize the error between the projected points $S_{2d}$ and the ground truth 2D corresponding positions $S_{2d,t}$:

$$\min_{f,\mathcal{R},\alpha_{id},\alpha_{exp},t_{3d}} \|S_{2d,t} - S_{2d}\| \quad (4)$$

According to the recent research on 3D Morphable Model fitting, we can just minimize the distance between the reconstructed model and ground truth on a set of sparse feature points. In contrast to 3D mesh points, the sparse set of feature points includes only dozens of points, such as contour, eyes, eyebrows, nose and other critical landmarks of the face. Owing to sparsity the fitting process converges quickly, without compromising accuracy of reconstruction provided the accuracy of 2D landmarks is adequate. Thanks to the recent advances in face alignment, 2D landmarking is now accurate, even in the case of large pose [18,19]. Thus, the fitting process on landmarks can be reduced to the following optimization problem:

$$\min_{f,\mathcal{R},\alpha_{id},\alpha_{exp},t_{3d}} \|S_{2d,t} - S_{2d}\|_{land} \quad (5)$$

## 4. Proposed Algorithm

In this section, we propose a novel method of Landmark Weighting for 3DMM Shape Fitting. The landmarks on the 2D image and the corresponding points on the 3D model have their own specific semantics, as shown in Fig.1. Therefore, the impact of fitting errors on different points varies. In order to understand this impact, it is necessary to analyze each pair of 2D-3D corresponding landmarks separately.

### 4.1. 3D Morphable Model Fitting with feature point weighting

Obviously, the reconstruction errors of the points in the corner of mouth and nose regions differ in expressions. For example, when a person is laughing or surprised, the shape of mouth changes more dramatically than that of the nose. Fig.2 illustrates the difference. However, the traditional Gauss-Newton method fitting process does not guarantee comparable matching accuracy for all points simultaneously, which may lead to the discrepancy in matching errors. For instance, the matching error of the mouth in the above case is greater than the nose matching error, and thus getting a bad reconstruction effect. To solve this problem, we propose a feature point weighted 3DMM fitting method. We add a matching weight to each landmark and bind the matching errors between these 2D-3D counterparts so that the errors of these points are balanced. The greater the distance between these corresponding points, the greater the weights. It can be expressed as follows:

$$\min_{\alpha,W,T,t_{3d}} \|W(S_{2d,t} - P(\overline{S} + S*\alpha + t_{3d}))\|_{land}^2 \quad (6)$$

where $W$ is the weighting matrix, $S_{2d,t}$ is the 2D ground truth input, $P = f\begin{bmatrix}1 & 0 & 0\\ 0 & 1 & 0\end{bmatrix}\mathcal{R}$ is the project matrix, $\overline{S}$ is the average shape, $S$ is the matrix of shape eigenvectors, $\alpha$ is the shape parameter and $t_{3d}$ is translation vector. We optimize these parameters using the ADMM method. Each group of parameters can be estimated while the other parameters are fixed. In order to constrain the reconstruction error of each point, we define a residual matrix $W = \text{diag}(w_1, w_2, \ldots, w_n)$, where $w_i$ represent the residual weight between $i$th 2D landmark and its corresponding 3D landmark projection.

### 4.2. Optimization

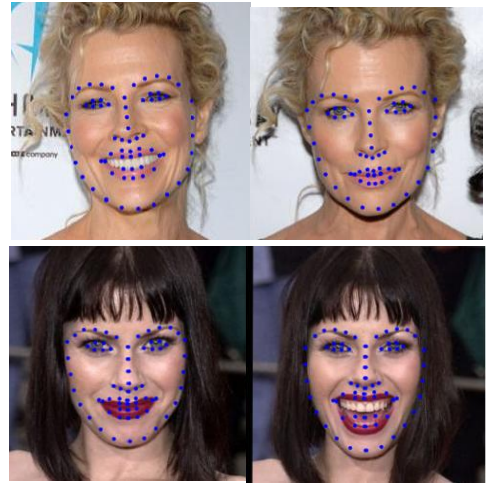

Figure 2. Different expressions with landmarks.

The key issue is how to determine matrix $W$. Our goal is to minimize the matching errors between these corresponding feature points and make these errors uniformly distributed at the same time, so as to avoid inferior fitting caused by excessive errors on some feature points. So, let us define vector $D$ as:

$$D = (S_{2d,t} - P(\overline{S} + S*\alpha + t_{3d})) \quad (7)$$

where $D \in \Re^{1\times n}$. We assume $d_i$ have a proportional relationship with weight $w_i$, the weight increases with the distance, i.e.

$$w_i = (d_i + \xi_1)/\xi_2 \quad (8)$$

where constant $\xi_1$, $\xi_2$ are used to ensure the stability of the weight and to prevent the weight being too large or too small. In this paper $\xi_1 = 3$, $\xi_2 = 3.5$, we determine their values by cross-validation. The weight $W = \text{diag}(w_1, w_2, \ldots, w_n)$ changes with error during each iteration.

In the optimization process, in order to prevent over-fitting, we need to add a regularization term in (6):

$$\min_{\alpha,W,T,t_{3d}} \|W(S_{2d,t} - P(\overline{S} + S*\alpha + t_{3d}))\|_{land}^2 + \lambda\alpha^T\mathcal{C}\alpha \quad (9)$$

where $\lambda$ is the coefficient for the regularization, the $\mathcal{C}$ is a constant matrix. $\mathcal{C} = \text{diag}(1/\sigma_1, 1/\sigma_2, \ldots, 1/\sigma_n,)$. $\sigma_i$ is the PCA standard deviation of 3DMM expression or shape, depending on whether the $\alpha$ is $\alpha_{exp}$ or $\alpha_{id}$. The closed-form solution is:

$$\alpha = \{(WPS)^T W(S_{2d,t} - P(\overline{S} + t_{3d}))\} / \{(\lambda\mathcal{C} + (WPS)^T WPS)\} \quad (10)$$

The optimization process of proposed method is summarized in Algorithm 1.

| Algorithm 1: 3DMM fitting with feature points weighting |
|---|
| **Data**: 2D landmarks ($S_{2d,t}$) of input image. |
| **Result**: 3D face mesh S. |
| 1  initialize landmarks weight matrix $W$. |
| 2  repeat |
| 3      estimate projections $f, \mathcal{R}, t_{3d}$ for each image; |
| 4      establish correspondence $P$ via back projection; |
| 5      estimate expression parameters $\alpha_{exp}$; |
| 6      estimate shape parameters $\alpha_{id}$; |
| 7      calculate expression and shape reconstruction error D = $\|S_{2d,t} - S_{2d}\|_{land}$ |
| 8      update weights $W$ via Eq.(7). |
| 9  until $\|S_{2d,t} - S_{2d}\|_{land} < \tau$ |

# 5. Experiments and analysis

## 5.1. Databases and experimental setup

In this section, we evaluate our method in two publicly available face databases AFW [32] and LFPW [33], which include labeled landmarks and a wide range of poses.

**AFW** dataset reference member (Zhu and Ramanan 2012) contains 468 faces in 205 images. Each face image is manually labeled with up to 6 landmarks and has a visibility label for each landmark.

**LFPW** dataset reference member (Peter N et al.2011) is a larger face dataset with 1432 faces downloaded from the web. Each image is labeled with up to 29 landmarks. We randomly select a subset of 300 images in the two datasets, and manually label additional landmarks up to 68 for all images.

The 3D Morphable Model we use in this paper is the Basel Face expression, so we add the face expression model to our 3DMM, from Face Warehouse [25]. It contains the neutral expression and 19 other expressions from 150 individuals spanning a wide range of ages and various ethnic groups. We use the $S_{id}$ = 199 bases of the Basel Face Model to represent identity variations and $S_{exp}$ = 29 bases for representing expression variations. In total, there are 228 bases representing 3D face shapes with 53215 vertices.

## 5.2. Evaluation metric

We compare the proposed method with E-3DMM [23]. We evaluate the reconstruction accuracy of fitting in different poses and expressions. The evaluation metric we used is Mean Euclidean Metric (*MEM*). In 3D reconstruction, our goal is to minimize the reconstruction error. MAE is defined as:

$$MEM = \sqrt{\frac{1}{N}\sum_{i=1}^{N}\sum_{j=1}^{k}\left\|(u_{ij}^g, v_{ij}^g) - (u_{ij}^r, v_{ij}^r)\right\|^2} \quad (11)$$

where N is the number of test samples, $k = 68$ is the number of landmarks of each face, $(u_{ij}^g, v_{ij}^g)$ and $(u_{ij}^r, v_{ij}^r)$ are the ground truth and estimated coordinates of the $j$th landmark of the $i$th sample. The distance we used in this paper is pixel distance.

## 5.3. Determining the weight

In order to find the optimal weights, an iterative optimization method has been used in this paper. The weights are automatically updated via Eq. (7) during the iteration. We iterated 50 times to optimize the weight in our experiment. The error is measured with MEM. In the experiment, we found that the error started converging after 20 iterations. Therefore, we chose the weight at 20$^{th}$ iteration as optimal value. Fig.3 shows the Mean Euclidean Metric error for shape(a) and expression(b) recovery at each iteration. As the updating of 3DMM parameters and weights alternates, the shape and expression reconstruction errors also decrease as the iterations increases.

## 5.4. Comparison Experiments

After determining the optimal weight, we compare the proposed method with E-3DMM. The authors collected large number of 3D scans with diverse expressions to train a shape model which can capture both facial and expression variations. As shown in Fig.4 and Fig.5, we carried out two sets of comparative experiments. We compared the proposed method with E-3DMM in shape and expression reconstruction. In a variety of angles, our method has significantly improved both in shape and expression reconstruction error, which can be seen in Fig.4 and Fig.5. For large pose, our method has more obvious improvement. Another observation is that the reconstruction error changes with pose. The greater the face rotation angle, the greater the reconstruction error.

It results from the fact that the rotation of locating landmarks in extreme poses is more difficulty.

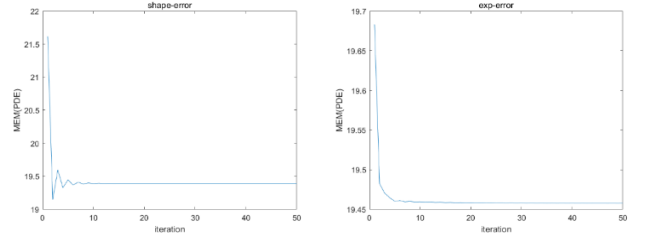

Figure.3. Iterative error curves for shape(a) and expression(b), respectively.

## 5.5 landmarks estimation

Our goal is not only to improve the overall reconstruction accuracy, but also to take the matching accuracy of each landmarks into account. Because of the different semantics of landmarks in different positions on the face of a person, the matching precision of these landmarks have different effects on the final reconstruction. We counted the estimation errors of landmarks using E-3DMM and FW-3DM respectively. There are four residual scatter plots in Fig.6, the horizontal axis represents the two coordinates of 68 landmarks, a total of 136 points. The vertical axis represents the residual of ground truth and the estimated position of the two coordinate axes. (a) and (b) denote the residuals distribution of the landmarks estimated using the traditional 3DMM fitting and proposed method respectively without considering expression. After adding expression component, the estimation errors of these two methods are shown in (c) and (d) respectively.

We can see that in Fig.6, comparing (a) with (b), the residuals of these landmarks are less scattered away from zero when residuals weight is taken into account. Not only the overall residuals are smaller, but also the residuals at each landmark tend to diminish. Note that larger residuals are reduced more significantly. According to our weighting rule, the effect of the feature points with large residuals is amplified. Therefore, the estimation errors on all feature points are more uniformly distributed. It avoids poor fitting results of the 3D face reconstruction.

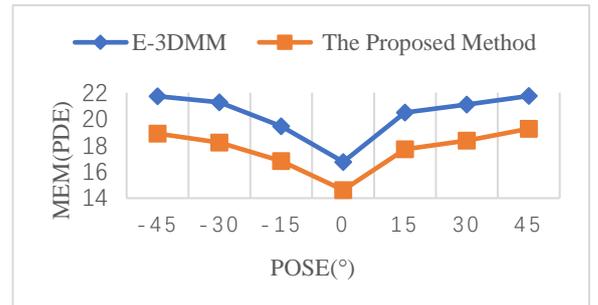

Figure 4. Shape MEM in different pose.

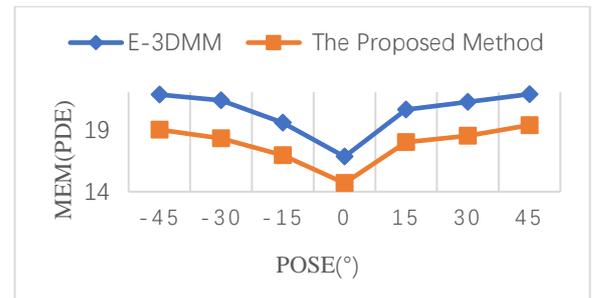

Figure.5. Expression MEM in different pose.

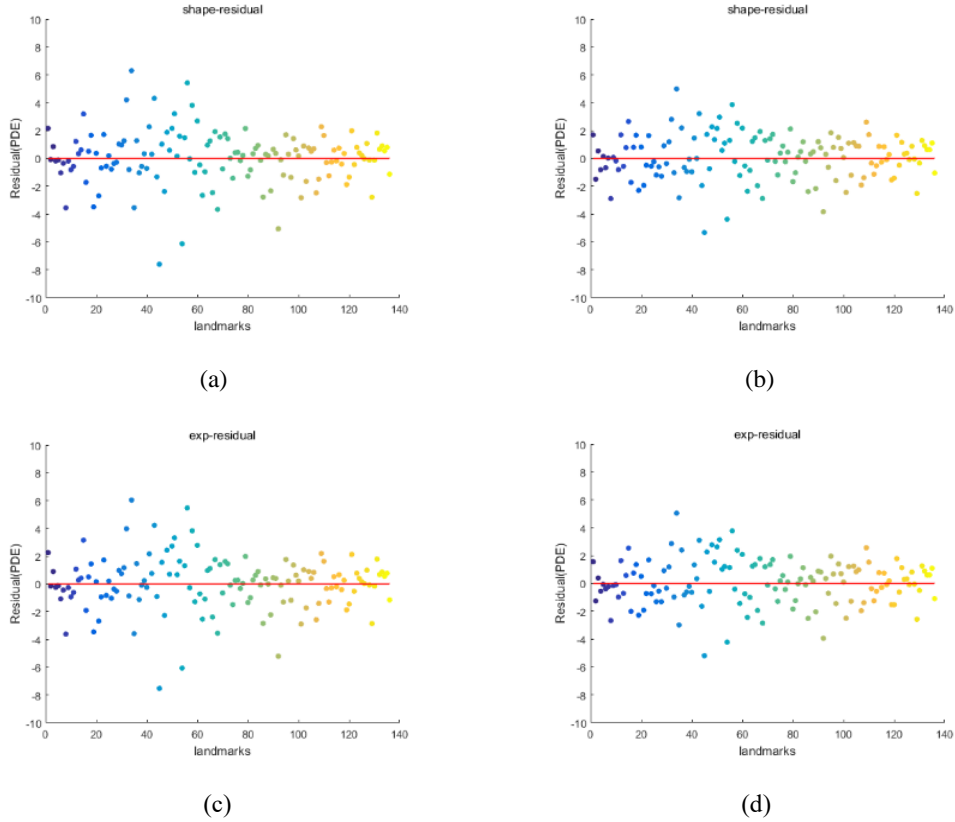

Figure.6. The reconstruction errors of shape and expression at each landmark point. (a) and (b) denote the residual distribution of landmarks on the expression after fitting by E-3DMM and proposed method, respectively. (c) and (d) represent the residuals of the two methods to estimate the landmarks in the shape, respectively.

Table.1 The shape reconstruction accuracy improvement in different pose.

| Dataset | -45° | -30° | -15° | 0° | +15° | +30° | +45° |
|---|---|---|---|---|---|---|---|
| AFW | 13% | 14.27% | 13.58% | 12.77% | 13.61% | 12.94% | 11.43% |
| LFPW | 10.89% | 11.38% | 10.37% | 11.29% | 11.51% | 11.93% | 11.21% |

Table.2 The expression reconstruction accuracy improvement in different poses.

| Dataset | -45° | -30° | -15° | 0° | +15° | +30° | +45° |
|---|---|---|---|---|---|---|---|
| AFW | 12.94% | 14.29% | 13.41% | 12.67% | 12.64% | 12.84% | 11.41% |
| LFPW | 10.69% | 14%. | 10.34% | 11.65% | 10.92% | 11.85% | 11.14% |

Table.1 shows the relative improvement of shape reconstruction accuracy of the proposed method at different angles, and Table.2 is the expression reconstruction improvement. From the experimental results, in these two representative face databases, our method achieves at least 10% improvement in accuracy at various poses.

Table.3 The fitting error for shape and expression at AFW.

| Method | shape | expression | sum |
|---|---|---|---|
| E-3DMM | 20.365 | 20.457 | 40.822 |
| FW-3DMM | **17.702** | **17.821** | **35.523** |

Table.4 The fitting error for shape and expression at LPFW.

| Method | shape | expression | sum |
|---|---|---|---|
| E-3DMM | 18.591 | 18.733 | 37.324 |
| FW-3DMM | **16.505** | **16.577** | **33.082** |

Table.3 and Table.4 present the reconstruction errors of E-3DMM and the proposed method obtained on the two databases, respectively. From the results we can see that our approach offers a significant improvement of 3D shape and expression reconstruction accuracy.

### 5.6 Fitting results

For the qualitative evaluation we selected images of subjects of different ages, gender, race different lighting, expression, pose from the AFW database. Fig.7 displays the fitting results by our algorithm. We can see from Fig.6 that in very complex conditions, the experimental results suggest that our approach is efficient and robust in complex environments.

Our algorithm can be used not only for the landmark-based 3DMM fitting, but also for any method that requires feature point-based fitting. For example, extracting SIFT or HoG features from facial images and using the 3DMM fitting based on these features can also be combined with the proposed algorithm to constrain feature points according to its semantic information or importance.

### 6 Conclusions

We have proposed a new fitting method to reconstruct a 3D face shape from a single 2D face image. The existing fitting methods do not take the importance of each point into account, but rather treat all points equally. However, every point on the face evidently has different semantics, and the reconstruction accuracy of each point has a different effect on the final fitting result. The reconstruction error of each landmark is modified by the proposed method, and the weight of each landmark adapted accordingly. The experimental results show that the total residual error distribution is more balanced as a result of the proposed weighting strategy. The estimation error of individual landmark is also reduced, and the global error becomes smaller. In future, we will study alternative ways to determine the weight matrix.

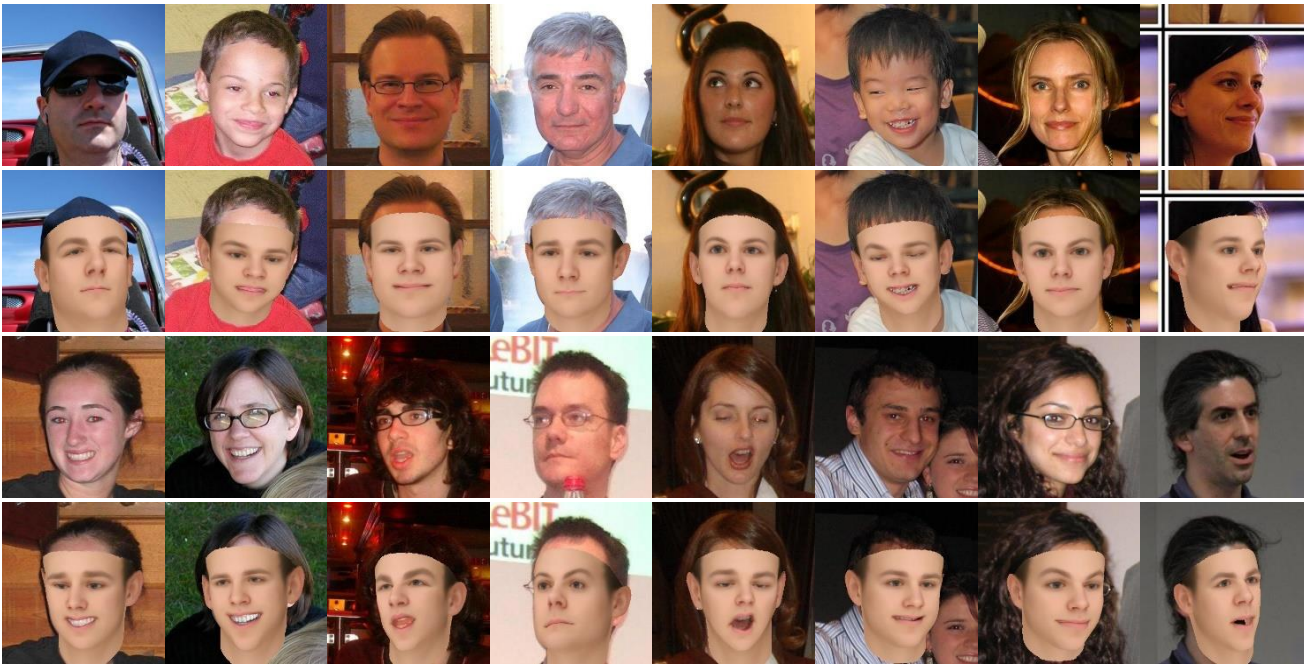

Figure.7. Reconstruction of 3D face for images in AFW by the proposed method.